%% file: ictai_pascaru.tex
\documentclass[conference]{IEEEtran}
\IEEEoverridecommandlockouts
\usepackage{cite}
\usepackage{amsmath,amssymb,amsfonts}
\usepackage{graphicx}
\usepackage{textcomp}
\usepackage{xcolor}
\usepackage{booktabs} 
\usepackage[ruled]{algorithm2e} 
\usepackage{todonotes}
\usepackage{csquotes} 
\usepackage{hyperref}
\usepackage{enumerate}
\usepackage{relsize}
\usepackage{array}

\hypersetup{
    colorlinks=true,
    linkcolor=blue,
    filecolor=magenta,      
    urlcolor=blue,
    citecolor=blue,
}

\def\BibTeX{{\rm B\kern-.05em{\sc i\kern-.025em b}\kern-.08em
    T\kern-.1667em\lower.7ex\hbox{E}\kern-.125emX}}
\begin{document}

\title{Vehicle routing and scheduling for regular mobile healthcare services}

\author{\IEEEauthorblockN{1\textsuperscript{st} Cosmin Pascaru}
\IEEEauthorblockA{\textit{Department of Computer Science} \\
\textit{Alexandru Ioan Cuza University of Iasi}\\
Iasi, Romania\\
cosmin.pascaru@info.uaic.ro}
\and
\IEEEauthorblockN{2\textsuperscript{nd} Paul Diac}
\IEEEauthorblockA{\textit{Department of Computer Science} \\
\textit{Alexandru Ioan Cuza University of Iasi}\\
Iasi, Romania\\
paul.diac@info.uaic.ro}
}

\maketitle

\begin{abstract}
We propose our solution to a particular practical problem in the domain of vehicle routing and scheduling. The generic task is finding the best allocation of the minimum number of \emph{mobile resources} that can provide periodical services in remote locations. These \emph{mobile resources} are based at a single central location. Specifications have been defined initially for a real-life application that is the starting point of an ongoing project. Particularly, the goal is to mitigate health problems in rural areas around a city in Romania. Medically equipped vans are programmed to start daily routes from county capital, provide a given number of examinations in townships within the county and return to the capital city in the same day. From the health care perspective, each van is equipped with an ultrasound scanner, and they are scheduled to investigate pregnant woman each trimester aiming to diagnose potential problems. The project is motivated by reports currently ranking Romania as the country with the highest infant mortality rate in the European Union.

We developed our solution in two phases: modeling of the most relevant parameters and data available for our goal and then design and implement an algorithm that provides an optimized solution. The most important metric of an output scheduling is the number of vans that are necessary to provide a given amount of examination time per township, followed by total travel time or fuel consumption, number of different routes, and others. Our solution implements two probabilistic algorithms out of which we chose the one that performs the best.
\end{abstract}
\vspace{0.1cm}

\input{motiv_intro}
\input{related_work}
\input{modeling}
\input{problem}

\input{heuristic_algorithm}

\input{genetic_algorithm}

\input{results}
\input{conclusion}

\bibliographystyle{IEEEtran}
\bibliography{ictai_pascaru}

\end{document}

%% file: motiv_intro.tex
\section{Introduction and Motivation}\label{Introduction and Motivation}

The initial motivation of the described work is the budget estimation for a project in the healthcare domain. The first concern is the initial cost that depends on the price and amount of required resources. Medical equipment is expensive, so the used metric is the number of vans with proper equipment. It mainly depends on the quantity of provided services, i.e., number of medical investigations, township configuration and route distances. A first phase was to gather statistics about how many patients lack the provided services in each location, based on previous years data.

Romania had an \emph{infant mortality rate} (IMR) of 7.0 in 2016 \footnote{ \href{http://appsso.eurostat.ec.europa.eu/nui/show.do?dataset=demo_minfind\&lang=en}{http://appsso.eurostat.ec.europa.eu/nui/show.do?dataset=demo\_minfind} }, as reported by the European Commission. IMR is an indicator that counts the number of deaths of children under one year of age per 1000 live births. This is the highest of all member states of the European Union. IMR is considered to be one of the most relevant indicators of a nation's health and social condition \cite{macdorman2009challenge}, and it is essential that a health care system implements strategies to lower it. One essential condition to do this is increasing the number of sonographies and prenatal tests that pregnant women make. Sonographies are recommended each trimester of pregnancy as they can reveal a large variety of fetus health issues that can be treated later.

The project is proposed for the Iasi county which is the most populated county in Romania after country capital Bucharest; that also has one of the highest infant mortality rates. If the project is successful, it is applicable in other counties as well. Destinations of interest are \emph{townships} around county capital, rural communities and do not have their own hospitals. We exclude other cities and townships very close to county capital since they have much easier access to hospitals and better medical services. There are \textbf{93} such townships, and the drive distance between any of them is under \textbf{three} hours. The project proposes the acquisition of a number of vans; provide each of them with sonographs, a diagnostic sonographer specialist, and a driver; and schedule them on daily routes. Each daily route of a van contains a list of visited townships in order, each assigned with some investigations planned for that day. Medical investigations for our purpose are planned to take \textbf{30} minutes. According to data provided by Romanian National Statistics Institute - INS \footnote{ \href{http://www.insse.ro/cms/en}{ http://www.insse.ro/cms/en \label{INS}} }, in 2016 in townships in Iasi county, \textbf{9.88\%} live births occurred without any prenatal testing, and another \textbf{12.43\%} had delayed their first testing to the second or even third trimesters. We exclude townships located less than 10 kilometers away from county capital.

The start and end points of any daily route are in county capital. We determined driving distances between any pair of destinations by calls to the Google Maps API \footnote{ \href{https://cloud.google.com/maps-platform/}{https://cloud.google.com/maps-platform/} } beforehand and saved as constant data for the scheduler. This method results in a reasonable approximation, as real-time traffic delays are harder to consider at this time of planning which is computed before any departure.

The project is part of a broader and nationwide initiative to improve general healthcare in disadvantaged areas in Romania. Considering the relatively poor infrastructure of rural areas, lack of medical offices and specialized personnel, this depends on using \textbf{mobile} medical units. For this, it was necessary to modify legislation to allow them and include proper methodological norms. A long time after the first proposal, this was finally approved by the Ministry of Health of Romania in May 2018 \footnote{ \href{http://www.cdep.ro/pls/legis/legis\_pck.htp\_act?ida=151171} {http://www.cdep.ro/pls/legis/legis\_pck.htp\_act?ida=151171} (Romanian) }.

The rest of the paper organizes as follows: Section \ref{Related Work} presents related work from two different perspectives: theoretically, general computer science or artificial intelligence domain similar problems; and more practical and specific healthcare applications or similar implemented projects and studies. Section \ref{Modeling} describes our modeling of the accessible data and how the solution is designed, as input data and requirements at output, and finally reduced to the formal optimization problem defined in Section \ref{Problem Definition}. The two proposed, and then compared algorithms used for both routing and scheduling are described in Sections \ref{Heuristic Algorithm} and \ref{Genetic Algorithm} with results in Section \ref{Results}. Finally, Section \ref{Conclusion and Future Work} concludes the paper and shortly discuses our expectations on possible requirement changes and maintenance work expected at the project implementation phase.

%% file: related_work.tex
\section{Related Work}\label{Related Work}

Most of the work described concentrates on initial design, available input data and requirements modeling, problem formulation and finally on the algorithm used for routing and scheduling. Almost all of it is applicable in other domains outside health care, for example, we can imagine for providing any sort of periodical maintenance or general tasks at given locations around a central node, that is the residency of working agents. However, the computer science problem as we formulate it in Section \ref{Problem Definition} is still surprisingly particular, and it is hard to find a previous similar study.

\subsection{Related Problems in Computer Science}\label{related_cs}
We fall under the categories of well-known problems: Traveling Salesman Problem - \textbf{TSP} and Vehicle Routing Problem - \textbf{VRP}. They contain a large number of variations, some of which share similarities to our problem. Multiple Traveling Salesperson Problem - \textbf{mTSP} is specifically resembling as we work with multiple vans. In \cite{bektas2006multiple} several variations of \textbf{mTSP} are presented, but only one category named \emph{Crew scheduling} mentions an application where the number of vehicles used is the optimization criteria. Almost all \textbf{mTSP} problems minimize the total travel distance or other costs when the number of vehicles is part of the input. They variate more on input specification, allowing restrictions of different types. Our version of \textbf{mTSP} is most particular on output specification, minimizing the number of vans. This is a vital particularity of the problem that we formulate, generated by the context we deal with in practice. Another important difference is that we allow multiple visits of the same node on a single route, with or without scheduling investigations during visits, a feature that seems more realistic.

Out of \textbf{VRP} problem variations, Periodic Vehicle Routing Problem - \textbf{PVRP} like the version studied in \cite{gaudioso1992heuristic} especially resembles our version, first because it requires allocation of periodic deliveries - that are medical services in our case, and minimizing the \emph{fleet size} - number of vans. There are still many differences though, as in \textbf{PVRP} there is no time spent on delivery, periodicity can differ on each customer, and vehicles have capacities. Neither of these is our case.

All of the problems described above and their variations are intractable and belong to the \textbf{NP-Hard} complexity class. That implies that no current polynomial deterministic algorithms are known that produce global optimum output. However, because the applications in practice are of high interest, many solutions were designed that provide sufficiently good outputs, or in some cases even proven to be very close to global optimum. Studied for a long time - more than just last decades, this solutions evolved in different directions of the same flavor. Such methods include metaheuristics like genetic algorithms \cite{grefenstette1985genetic}, tabu search \cite{malek1989serial}, simulated annealing, other bio-inspired optimizations \cite{necula2017tackling}. Other types of approaches are machine learning \cite{gambardella1995ant} or distributed algorithms \cite{pavone2011adaptive}, \cite{ozden2017solving}.

\subsection{Related Work in Healthcare}
Many studies that highlight improvements of infant (or neonatal, perinatal or maternal) mortality rates by increasing accessibility to prenatal testing especially providing obstetric ultrasonography. Even more numerous benefits of prenatal medical investigations can be an earlier observation of less severe abnormalities, that can be far from life-threatening or easily treatable. However, it takes too much effort to collect statistics about many of these, and for now, we decided to evaluate impact based only on the most important and easy to obtain indicators. An experimental project has been conducted in rural areas in Uganda \cite{ross2013low}, with excellent results; which might be expected because of much higher infant mortality rates in Uganda. Other interesting studies are presented in \cite{kongnyuy2007use}, \cite{traore2013use} or \cite{iams2008primary}. Also, recently discovered technologies could greatly reduce the cost of medical equipment with sufficient accuracy of scans, as shown in \cite{van2017development}.

To know about specific local issues that arise in implementing our project we had several meetings with medics that founded \emph{Medics Caravan} \footnote{ \href{https://www.caravanacumedici.ro/en/home.html}{https://www.caravanacumedici.ro/en/home.html} }. This project started in 2014 as a nongovernmental organization with the aim to provide general essential medical services in isolated rural areas in Romania. Their activity is less interesting from a computer science perspective, as they only schedule occasional visits. However, it helped us set our expectations towards more practical difficulties, like how to announce a visit to a remote village or if it is feasible to rely on local electric power sources.

%% file: modeling.tex
\section{Modeling}\label{Modeling}
The initial user requirements are to determine the minimum number of vans necessary for satisfying a number of medical examinations that suffice all pregnancies. On top of this, we also computed the routes themselves, and later added the scheduling of vans by days. We decide the number of examinations at the very first step. Two reasons motivate the choice of working with townships. First, their suitable number of \textbf{93} is small enough to allow fast computations described in algorithms \ref{Heuristic Algorithm} or \ref{Genetic Algorithm} (as opposed to the larger number of \textbf{331} villages) and second, for every village it is close enough to reach the center of a township.

Data provided by INS \textsuperscript{\ref{INS}} was for each year from 2000 to 2016, and for each township containing: the total number of live births, out of which how many had no prenatal examination, and that had their first examination in first, second and third trimesters of pregnancy. We considered only the numbers from the most recent year and summed up births without any examination or with first examinations in second or third trimesters. For each township, this number is multiplied by seven as it is the recommended number of examinations for any pregnancy. We decided to have the scheduling for a month so the numbers obtained were divided by twelve, rounding to the higher integer. The assumption is that all months follow the same schedule, only patients change and that births are evenly distributed by months. Also, we considered \textbf{21} working days in a month as it is the average and a limit of \textbf{10} working hours a day.

A route takes all day and is specified by an ordered list of townships each with the associated number of examinations. The total time is the sum of driving times plus \textbf{30} minutes for each examination on the route. Under this model, the challenge is to find the most efficient (i.e., smallest) set of routes that sum up all examinations of each township. Then the number of necessary vans is the number of routes divided by \textbf{21}, rounded to the higher integer. Since the vans are homogeneous, it is not important what vans cover what route at this point, so they are just assigned to a different route each day.

%% file: problem.tex
\section{Problem Definition}\label{Problem Definition}
\subsection{Parameters} There are a series of constant parameters that we assigned values to based on practical considerations. These are \textbf{30} minutes as the duration of examinations and \textbf{10} working hours of each of \textbf{21} working days in a month. In our implementation, they are easy to change by modifying a configuration file.

\subsection{Input}
Problem inputs consist of the township configurations, distances and necessary examinations that is the population of interest. We precomputed a $(N+1)\times(N+1)$ matrix of all distances between all pairs of the $N = $ \textbf{93} townships plus the capital. The values are integer numbers representing driving minutes. We also kept a mapping from numeric ids, that the core algorithms work with, to the string names used at output and number of monthly examinations of each township. Let:
\begin{itemize}
 \item $E_{i} \Leftarrow $ number of examinations necessary in township $i$
 \vspace{5pt}
 \item $ { } D_{i, j} \Leftarrow \hspace{4pt}_{\text{\normalsize{includes capital conventionally at index 0}}}^{\text{\normalsize{travel time distance between locations $i$ and $j$,}}} $
\end{itemize}
\subsection{Output}
The output consists of a list of scheduled tours, such that all townships have the specified number of examinations. Let:
\begin{itemize}
 \item $T_{i} \Leftarrow $ list of tours, $1 \leqslant i \leqslant $ M (number of tours)
 \vspace{5pt}
 \item $T_{i,j} \Leftarrow \Big\{ \hspace{4pt}_{\text{\normalsize{$1 \leqslant j \leqslant $ $\vert T_{i} \vert \hspace{4pt} \big( $tour length\big)}}}^{\text{\normalsize{$j^{th}$ township id of tour $i$,}}}  $ 
  \vspace{5pt}
  \item $S_{i,j} \Leftarrow \Big\{ \hspace{4pt}_{\text{\normalsize{$1 \leqslant j \leqslant $ $\vert T_{i} \vert \hspace{4pt} $}}}^{\text{\normalsize{examinations scheduled at $T_{i,j}$ in tour $i$}}}  $ 
  \vspace{5pt}
\end{itemize}
One solution is valid first it satisfies all required examinations:
\begin{displaymath} \mathlarger{E_{t}} \hspace{10pt} = \hspace{-20pt} \sum_{\substack{(i,j)\in(\{1..M\} \times \{1..\vert T_{i} \vert\})\\ \text{\normalsize{with }} \mathlarger{T_{i,j}=t}}} \hspace{-6pt} S_{i,j} \text{\normalsize{ , for any tour $1 \leqslant t \leqslant $ M}} \end{displaymath}
And if any scheduled tour $1 \leqslant t \leqslant $ M has a total duration of at most \textbf{10} hours:
\begin{displaymath} \sum_{\substack{i = 2}}^{\vert T_{t} \vert}{D_{T_{(t,i-1)},T_{(t,i)}}} + D_{0,T_{(t,1)}} + D_{T_{\vert T_{t} \vert},0} + 30 \times \sum_{\substack{i = 1}}^{\vert T_{t} \vert}{S_{t,i}} \leqslant 600 \end{displaymath}

\subsection{Optimization criteria}
One solution $(T\emph{1},S\emph{1})$ is better than another $(T\emph{2},S\emph{2})$ if $\vert T\emph{1} \vert < \vert T\emph{2} \vert$. If $\vert T\emph{1} \vert = \vert T\emph{2} \vert$, then $(T\emph{1},S\emph{1})$ is better iff the sum of the total duration of all tours in $T\emph{1}$ is smaller than the one of $T\emph{2}$. The sum of $S$ - examination time is irrelevant in comparing because it is constant. Other optimizations criteria can be considered, for example like reducing work hours imbalance by days or using a smaller number of distinct \emph{routes} but for now, these are left for future work.

\subsection{Particularities relative to classic problems}
Similar problems mentioned in section \ref{related_cs} are well-known in literature,  each with its particularities. Likewise, under the definition above, our problem has the following particularities:

\begin{itemize}
 \item \textbf{minimum number of vans}/vehicles/salesman or agents, in general, is the most important optimization criteria. This is far less frequently meet in other papers.
 \item \textbf{multiple visits of each township}: nodes have a number of similar \emph{tasks} that have to be done by agents, which consume time. Passing through a node multiple times is allowed even without doing any task.
 \item \textbf{time constraints on tour length}: all routes represent working days that are limited to a number of hours. This is easier to handle than the more frequent \emph{time windows constraints}, and in our case reasonable for patients, as eventually, time preferences of patients can be satisfied by a different scheduler algorithm.
  \item \textbf{all tours start and end in the capital}, known as the \emph{single depot} in vehicle routing.
\end{itemize}

The examinations periodicity in initial user requirements was solved by the way the problem was modeled. The number of examinations per month is an estimate and being part of input it is easy to change, which will probably happen often during project implementation. Similar the year data, months and the \textbf{21} working days of a month are part of problem design. In particular though, it was relevant to reach a minimum number of tours but divided by \textbf{21}, rounded to the higher integer as this gives the number of necessary vans.

%% file: heuristic_algorithm.tex
\section{Heuristic Algorithm}\label{Heuristic Algorithm}

Like all solutions of related problems presented in \ref{related_cs}, both our approaches are approximation algorithms. The first reason is that there are no known algorithms that produce optimum solutions (under the assumption that the problem is \textbf{NP-Hard} as all similar problems). Second, by the way we designed our heuristics, it is easy to adapt to relative small problem changes like changing the optimization criteria or even solution structure.

The heuristic algorithm described in this section provides the best results on the given dataset. The main structure of the algorithm came when we tried to devise a simple mechanism to make sure the vans follow relatively \emph{good} paths, and that proved to be more than enough for a very good solution. 

The algorithm contains three parts:
\begin{enumerate}[\itshape(A.)]
  \item generating a relatively large number of small tours that the vans could follow. 
  \item create the complete scheduling by assigning vans to the tours, and performing a certain number of examinations in every visited township.
  \item score a given planning based on certain parameters.
\end{enumerate}

\subsection{Tour generation}
We want the smaller tours to be \emph{optimal} with respect to the number and order of visited townships. Also, it is preferred to be easy to select a subset that can cover all townships, but every tour should also be short enough to be covered in one working day, including potential examinations. This can be seen as a relaxation of the \textbf{mTSP}, as we allow overlapping tours. However, given that we do not know the distribution of examinations at this point in the algorithm, we generate multiple solutions to a \textbf{mTSP} problem and consider all the relevant small tours that have emerged.

As such, any solver for the \textbf{mTSP} problem can be used. It is not always necessary that better \textbf{mTSP} solutions lead to an improvement in the overall result, as sometimes more variance in the small tours is more important than the \textbf{mTSP} global optimum that they generate.

Our solution uses a simple simulated annealing \cite{kirkpatrick1983optimization} procedure to generate the \textbf{mTSP} solutions.

Simulated annealing is a meta-heuristic successfully used in many applications for finding (or rather, approximating) the global optimum of a function. It starts with an initial random solution, and, at each time step, it generates a neighbor solution, evaluates it, and decides to move to it or not based on which one is better, and by how much. The algorithm first has a high probability of accepting worse solutions, that gradually decreases over time; this is designed to simulate the process of annealing in metallurgy, hence the name of the algorithm.

\subsection{Planning generation}

As input, there are: the set of available tours that we can use, and the required number of examinations that have to be done in every township.

As long as there are still unvisited townships, we choose a tour together with examinations that should be done on that tour. Then we remove those from the required remaining examinations of corresponding townships.

The choice of the tour is made through a statistical selection. A score is computed for the choice of every possible tour, based on the number of examinations that can be done in that tour (the computation of which can be seen in Algorithm \ref{alg:compute_examinations}), their importance, and the total traveled distance. The entire scoring mechanism is presented in Algorithm \ref{alg:compute_tour_score}. Then a tour is selected with a weighted random from the top x\% (every tour has a probability of being selected directly proportional to its score). This is described in Algorithm \ref{alg:choose_tour}.

The choice of what townships should have scheduled examinations during a tour could be done through multiple strategies: 

\begin{itemize}
	\item closest first
	\item furthest first
	\item most relevant first
	\item randomly
\end{itemize}

The strategies also incorporate a mechanism to limit the number of \emph{single examinations} (traveling to a township, especially a faraway one and only being able to do one or two examinations). This situation can happen because there are few remaining examinations to be done in that specific township, or the remaining time during the working day prevents a longer stay. The entire strategy for generating a complete planning is described in Algorithm \ref{alg:generate_planning}.

\begin{algorithm}[h]
	\caption{Compute Examinations \label{alg:compute_examinations} }
	\KwIn{\textbf{t}, the tour, and \textbf{E[]}, the required number of examinations per township}
	\KwOut{An array of the same length as the input tour, where for every element, the value represents the number of examinations to be done in that certain township.}
	Sort all townships in current tour, furthest from origin come first\;
	Set the array of examinations to be done as empty\;
	\While{the time required to go through the currently selected visits is still smaller than the maximum working time per day}{
		Select the next township and, if the remaining time allows it, add one examination to it\;
		Add as many examinations as possible to this specific township, until all necessary ones are completed, or there is not enough time remaining\;
	}
	Return the computed array of examinations\;
\end{algorithm}

\begin{algorithm}[h]
	\caption{Compute Tour Score \label{alg:compute_tour_score} }
	\KwIn{\textbf{t}, the tour, and \textbf{E[]}, the required number of examinations per township}
	\KwOut{A single float value, the score of the tour.}
	Compute the examinations to be done by following tour \textbf{t}\;
	Return the score as the total number of examinations, divided by the total distance (from a greedy selection perspective, this maximizes examinations done per distance unit)\;
\end{algorithm}

genetic algorithms
\begin{algorithm}[h]
	\caption{Choose Tour}
	\label{alg:choose_tour}
	\KwIn{\textbf{T[]}, the list of available tours, and \textbf{E[]}, the required number of examinations per township}
	\KwOut{A single tour from \textbf{T}, to be selected at the current step}
	Compute scores for each tour\;
	Sort tours descending by the score\;
	Keep only top 10\% tours\;
	Choose randomly from remaining tours, with probability proportional to score\;
\end{algorithm}

\begin{algorithm}[h]
	\caption{Generate Planning \label{alg:generate_planning} } 
	\KwIn{\textbf{T[]}, the list of available tours, and \textbf{E[]}, the required number of examinations per township}
	\KwOut{\textbf{P[]}, a planning, each element consisting of a tour and examinations per tour}
	\textbf{P} = []\;
	\While{ $\exists \hspace{2pt} \textbf{E[i]} \neq 0$ }{	
		Choose \textbf{t} as the tour to follow at the current step\;
		Compute where and how many examinations should be done, throughout the tour \textbf{t}\;
		Remove all new examinations from the remaining ones\;
		Add tour \textbf{t} and the associated examinations to the output planing \textbf{P}\;
	}
	\Return{P}
\end{algorithm}

\subsection{The score of a scheduling}
Computes the score of a full schedule, based primarily on the number of tours that have to be done, but also on total distance traveled (which translates to fuel consumption) and the number of distinct tours used. By minimizing the number of distinct tours, the drivers can learn them by heart, simplifying the real-world application of the entire process. This score is computed as a weighted sum of all those factors, and the relevance of each one is parametrized.

%% file: genetic_algorithm.tex
\section{Genetic Algorithm}\label{Genetic Algorithm}


We implementing a second algorithm to have other solutions to compare with, i.e. see if the problem instance is solvable by a smaller number of routes. For this, we choose a genetic algorithm as they are known for solving similar problems \cite{grefenstette1985genetic}.

A genetic algorithm is a metaheuristic inspired by the natural process of evolution of species. A population of individuals also known as chromosomes, each one representing a solution is evolved over multiple generations. 

At each generation, the new population is selected from the old one, after some individuals undergo \textbf{crossover}, that exchange parts of the solution representation, and \textbf{mutations} that randomly change the underlying solution. The chromosomes with a higher score have a higher probability of being selected for the next generation, although lower score solutions are necessary for preserving the variance of the population, especially during the later generations. 

\subsection{Complete or ideal representation of a solution}

A complete solution to the presented problem would have to encode a list of tours that the vehicles have to follow, and the number of examinations that have to be performed at each township along the way. This representation is quite complex for a genetic algorithm to manage, as the required operators, especially the crossover operator, would have to be very complicated to preserve the validity of the solutions. Because of this, we propose the following alternative representation.

\subsection{Real representation used of an individual}

In our implementation, an entire solution is represented using only a single permutation of the numbers from 1 to the total number of examinations.

In the classical \textbf{mTSP} problem, each location must be visited by a salesman exactly once. To solve a more similar problem, the input is transformed, by duplicating the locations that we need to visit multiple times, instead of having to count the number of examinations in each one. As such, we can represent an order of solving every visit by such a permutation.

This representation would be complete if the goal were to have only one vehicle go through all the required visits. However, we need a strategy to split the permutation into multiple tours. 

Let $P$ be the permutation defined above, and $h$ be the county capital (the start and end point for every tour). The simplest way would be to assign, for each tour $T_i$, a sequence $P_l, P_{l+1}, ..., P_r$ in the permutation, such that the tour performs examinations in the following order: $h, P_l, P_{l+1}, ..., P_r, h$. Note that there will be no examinations performed in $h$ and that the locations $P_i$, although distinct examinations, can refer to the same township, as defined previously.

This assignment could be a part of the solution, evolved together with the permutation, or computed in a greedy fashion each time it is required. Both of the solutions are implemented, and the second proves to give better results, although it takes longer to achieve an average solution. From now on, we only consider the greedy approach.

This method is very simple, and defined as follows:

\begin{algorithm}[h]
	\caption{Generate Tour Assignment}
	\KwIn{\textbf{P[]}, the permutation of all examinations}
	\KwOut{\textbf{T[]}, a list of tours that result from permutation \textbf{P}}
	Let \textbf{T} be empty\;
	Set the current tour as empty\;
	\ForEach{examination in \textbf{P}}{
		\uIf {there is enough time remaining for examination to be added to the current tour}{
			Add examination to current tour\;
		}
		\Else{
			Add current tour to \textbf{T}\;
			Reset current tour\;
			Add examination to current tour\;
		}
	}
	\If{the current tour is not empty}{
		Add current tour to \textbf{T}\;
	}
	\Return{T}
\end{algorithm}

What follows is the proof that this greedy method of separating tours from the permutation cannot generate more tours than the optimal solution. It is possible that an assignment with a shorter total distance exists, but remember that the main objective is to minimize the total number of tours.

Let us assume that the permutation $P$ is divided by the greedy algorithm in $k$ tours, delimited by indexes $g_0, g_1, g_2, ..., g_k$. Conventionally for simplicity  $g_0$ and $g_k$ are the first, respectively last indexes of the permutation. Consider that the tours are: $\{(P_{g_0 + 1}, P_{g_0 + 2}, ..., P_{g_1})$, $(P_{g_1 + 1}, P_{g_1 + 2}, ..., P_{g_2})$, ..., $(P_{g_{k-1}+1}, P_{g_{k-1} + 2}, ..., P_{g_k})\}$. 

Let the optimal delimitation split the permutation into $h$ tours, where $h < k$. Denote this "best" delimitation as $b$, such that the tours are as follows: $\{(P_{b_0 + 1}, P_{b_0 + 2}, ..., P_{b_1})$, $(P_{b_1 + 1}, P_{b_1 + 2}, ..., P_{b_2})$, ..., $(P_{b_{h-1}+1}, P_{b_{h-1} + 2}, ..., P_{b_h})\}$. Similarly, $b_0$ and $b_h$ are the first, respectively last indexes of the permutation.

Let $w$ be the smallest value for which $b_w > g_w$, $w \in \{1..h-1\}$. This implies $\forall i \in \{0..w-1\}, b_i \leq g_i$. For sure, such a $w$ exists because $h < k$.
Given that $b_{w-1} \leq g_{w-1}$ and $b_w > q_w$ the following two cases can be distinguished.

\begin{enumerate}
	\item $b_{w-1} = g_{w-1}$: the greedy choice ensures that $g_w$ is the highest value for an index in the permutation that will give a valid tour that starts with examination $P_{g_{w-1}}$. This implies that adding another examination will generate an invalid tour, and so, the "optimal" solution would be invalid.
	\item $b_{w-1} < g_{w-1}$: in this case the $w$-th tour of the optimal solution will contain following examinations: $P_{b_{w-1}}$, $P_{b_{w-1}+1}$, ..., $P_{g_{w-1}}$, $P_{g_{w-1}+1}$, ..., $P_{g_{w+1}-1}$, $P_{g_{w+1}}$ ..., $P_{b_w}$. It is obvious that this tour is the same as the $w$-th tour in the greedy, with some locations added before and after. The ones added as prefix might generate a valid tour, but the ones added after making this case reduce to the first one: it's impossible to extend the greedy tour any further.
\end{enumerate}

Cases 1) and 2) conclude that the assumption is false, which implies that the greedy algorithm generates a solution that has at most as many tours as the optimal solution (given the input permutation).

\subsection{Mutation of a chromosome}

Three types of mutation are being used:

\begin{itemize}
	\item swap two random positions from the permutation
	\item reverse a random subsequence from the permutation
	\item shuffle a random subsequence from the permutation
\end{itemize}

The mutation that should be used each time is chosen randomly based on a distribution given as parameter. 

\subsection{Crossover of two chromosomes}


There are three crossovers being used: 
\begin{itemize}
	\item Ordered Crossover (OX) \cite{davis1991handbook}
	\item Partially Matched Crossover (PMX) \cite{goldberg1985alleles}
	\item Uniform Partially Matched Crossover (UPMX) \cite{cicirello2000modeling}
\end{itemize}

Similarly to how the mutations are chosen, each time a crossover is required, it is sampled from a random distribution.

\subsection{Evaluation of a chromosome}

The evaluation is straightforward; once the tours are selected from the permutation in the aforementioned greedy manner, the score of a chromosome is given by the number of tours, multiplied by a factor, plus the total distance traveled. Obviously, we want the factor to be large enough such that a solution with fewer tours but higher total distance will always be preferred to one with more tours. This forces the optimization of the number of vehicles to come first, as they are the most expensive resource.

%% file: results.tex
\section{Results}\label{Results}

\subsection{Statistic results}

Table \ref{tab1} displays the results of the two algorithms presented above on our dataset. All the results, except for the last line, are computed as the average after five runs.

\newcolumntype{P}[1]{>{\centering\arraybackslash}p{#1}}

\begin{table}[htbp]
\setlength\extrarowheight{2.5pt}
\caption{Average runs results}
\begin{center}
\begin{tabular}{|P{0.17\textwidth-2\tabcolsep - 3\arrayrulewidth}
                |P{0.1\textwidth-2\tabcolsep - 3\arrayrulewidth}
                |P{0.1\textwidth-2\tabcolsep - 3\arrayrulewidth}
                |P{0.13\textwidth-2\tabcolsep - 3\arrayrulewidth}|}
\hline
\textbf{Algorithm variation used} & \textbf{Time limit in seconds} & \textbf{Average number of tours} & \textbf{Average total distance} \\
\hline
		1) default heuristic algorithm & 20 & \textbf{42.2} & 369905.2 \\
		\hline
		2) closest locations first & 20 & \textbf{42.8} & 393990.4\\ 
		\hline
		3) different tour score computation & 20 & \textbf{44.0} & 385327.4\\
		\hline
		4) larger keep percent & 20 & \textbf{42.2} & 379067.2\\
		\hline
		4) larger keep percent & 40 & \textbf{42.0} & 377200.2\\
		\hline
		5) default genetic algorithm & 60 & \textbf{82.6} & 2667210.2\\
		\hline
		5) default genetic algorithm & 3600 & \textbf{63} & 565800\\
		\hline
\end{tabular}
\label{tab1}
\end{center}
\end{table}

Descriptions of algorithm versions used:
\begin{enumerate}
	\item The \textbf{default heuristic algorithm} uses the following parameters:
	\begin{itemize}
		\item selects the furthest locations first when choosing the townships to visit in every tour, as presented in Algorithm \ref{alg:compute_examinations} 
		\item the score of tours is computed as described in Algorithm \ref{alg:compute_tour_score}, that is the ratio between the number of examinations made and the total distance traveled; 
		\item the \emph{keep percent} is set to 20\% (how many top tours should be considered relevant when scoring them); this parameter is used in Algorithm \ref{alg:choose_tour}
	\end{itemize}
	\item The \textbf{closest locations first} algorithm drastically changes the order in which examinations are selected along a tour. This seems to give worse results, due to the fact that there are multiple faraway examinations to be made in the last days, and they are not grouped well, as they would have been in they were done earlier during the algorithm.
	\item The \textbf{different tour score computation} algorithm uses subtraction instead of ratio; basically computes the score as the difference between distance and number of examinations done (the latter being multiplied by a factor, in order to preserve relevance). In some situations, this alternate method proved to give better results.
	\item The \textbf{larger keep percent} algorithm increases the \emph{keep percent} to 40\% (up from 20\%), thus improving the search space of the algorithm, although increasing the running time for finding a good solution. By making this value even larger the algorithm would theoretically find even better solutions, although, in practice, because of the randomness and the time restrictions, it is not feasible.
	\item The \textbf{default genetic algorithm} uses a $\mu + \lambda$ strategy \cite{ter2011convergence}, with $\mu = 150$ (population size), $\lambda = 300$, a crossover probability of $0.6$ and a mutation probability of $0.2$. These parameters were chosen empirically after multiple runs. The operators used in the algorithm are described in Section \ref{Genetic Algorithm}
\end{enumerate}

\subsection{Visualization of routes and scheduling}
A method to visualize the resulting tours was also implemented. The scheduling is provided in simple text format, for each of the \textbf{21} days tours are printed for all vans, with the basic tour numbers and: total driving time, examination time, their sum giving total work time, and townships in order with the number of scheduled tours in each. From this, it is easy to produce a schedule for each township separately to see on what days and hours patients have appointments. Patients schedule preferences is out of our focus now and would probably be hard to collect this information.

\begin{figure}[h]
\centering
\includegraphics[width=0.5\textwidth]{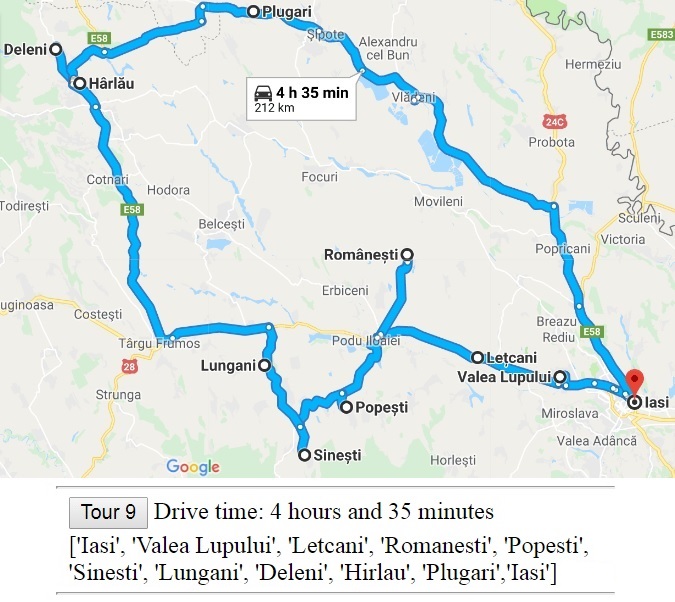}
\vspace{-13pt}
\caption{An example visualization of a basic tour}
\label{fig:tour_9}
\end{figure}

Tours are of interest especially for drivers. All basic tours (tours generated in the first part of the heuristic algorithm) are included in a web page, where each tour has an associated button loading a Google Maps frame displaying that tour, as shown in Fig. \ref{fig:tour_9}.

\vspace{-2pt}

%% file: conclusion.tex
\section{Conclusion and Future Work}\label{Conclusion and Future Work}

There are two future work directions:
\begin{itemize}
	\item improving algorithm performance, in speed and in generated solutions, also improve the general usability, adapting it to other real-life scenarios that might come up during the implementation of the project.
	\item actually starting the project, acquiring the vehicles, equipment, employing medic specialists and drivers; currently blocked by funding application.
\end{itemize}

Regarding the algorithm, a simple, but very promising way to improve the quality of the solutions would be to generate multiple good, but diverse solutions using the heuristic algorithm, and then pass them as input to the genetic algorithm. Given enough variance, it can greatly improve already very good solutions.

However, as the generated solutions prove that two vehicles are enough for our input data, there is not much to be gained in the real world application on input data as it is.

As for project implementation, the next phase is obtaining funding for the initial acquisition of at least two equipped vans, and find drivers and medic specialists. The implemented application is not blocking the project road-map currently. For now, it may be more important to change how the problem is modeled, what practical aspects are translated into input data and what are the most appropriate optimization criteria. It is not hard to adapt our application to many such changes but is harder to imagine the right ones before the project reaches the implementation phase. After real drivers and medics start to cover scheduled routes we expected that they will provide much feedback and request for new features, and prove that tours are doable in the times we assign them to or the contrary. Simple changes can be done by modifying parameters, score functions and even the returned solution structure to some extent.
More elaborate changes would be required if we would tackle the following, possibly useful aspects: managing data of pregnancy stages per patient level, managing appointments per patient with the possibility of rescheduling on missing appointments, including smaller settlements than townships or real-time re-scheduling when unexpected events are encountered in the field.

In conclusion, we believe that we reached the goal of finding the smallest number of vans that are needed initially for developing the project, and we also are able to generate and visualize the tours and scheduled appointments in each day of the month. Our application of this heuristic algorithm especially has shown that a more complex problem than a human can solve is easy to tackle by implementing such solutions. Moreover, it is now easy to reschedule the entire planning if any configurations change, apply the same algorithm to different regions as well, or adapt it to include more input elements or restrictions.

\section*{Acknowledgments}

We want to thank our colleague Dr. Cosette Chichirau for helping with problem specification and obtaining data from INS \textsuperscript{\ref{INS}}, Prof. Dr. Mihaela Breaban and Dr. Raluca Necula for help on genetic and heuristic algorithms and Dr. Mihai Ranete and his colleagues from \emph{Medics Caravan} team for sharing their experience in the field and providing feedback.